\documentclass[sigconf]{acmart}

\usepackage{xcolor,colortbl}
\usepackage{epsfig}
\usepackage{graphicx}
\usepackage{amsmath}
\usepackage{enumitem}

\usepackage[font=footnotesize,labelfont=bf]{caption}

\usepackage{multirow}
\usepackage{tabularx}
\usepackage{dsfont}
\usepackage{url}
\usepackage[tight]{subfigure}
\usepackage{subfigure}
\usepackage{amsthm}
\usepackage[export]{adjustbox}
\usepackage{comment}

\usepackage[utf8]{inputenc} 
\usepackage[T1]{fontenc}    
\usepackage{url}            
\usepackage{booktabs}       
\usepackage{amsfonts}       
\usepackage{nicefrac}       
\usepackage{microtype}      
\usepackage{pifont}
\usepackage{textcomp}
\usepackage{tikz}
\usepackage{sidecap}

\definecolor{dg}{rgb}{0,0.694,0.298}
\definecolor{purple}{rgb}{0.4,0.176,0.569}
\definecolor{tabgray}{rgb}{0.85,0.85,0.85}

\usepackage{pifont}
\newcommand{\figref}[1]{Fig.~\ref{#1}}
\newcommand{\reqref}[1]{Eq.~\eqref{#1}}
\newcommand{\secref}[1]{Sec.~\ref{#1}}
\newcommand{\tableref}[1]{Table~\ref{#1}}

\usepackage{xspace}
\makeatletter
\DeclareRobustCommand\onedot{\futurelet\@let@token\@onedot}
\def\@onedot{\ifx\@let@token.\else.\null\fi\xspace}
\def\eg{\emph{e.g}\onedot} 
\def\ie{\emph{i.e}\onedot} 
 
\def\etc{\emph{etc}\onedot} 
 
\def\etal{\emph{et al}\onedot}
\makeatother

\AtBeginDocument{%
  \providecommand\BibTeX{{%
    \normalfont B\kern-0.5em{\scshape i\kern-0.25em b}\kern-0.8em\TeX}}}

\setcopyright{acmcopyright}
\copyrightyear{2021} 
\acmYear{2021} 
\setcopyright{acmcopyright}\acmConference[MM '21]{Proceedings of the 29th ACM International Conference on Multimedia}{October 20--24, 2021}{Virtual Event, China}
\acmBooktitle{Proceedings of the 29th ACM International Conference on Multimedia (MM '21), October 20--24, 2021, Virtual Event, China}
\acmPrice{15.00}
\acmDOI{10.1145/3474085.3475170}
\acmISBN{978-1-4503-8651-7/21/10}

\begin{document}
\fancyhead{} 

\title{\emph{JPGNet}: Joint Predictive Filtering and Generative Network \\ 
for Image Inpainting}

\author{Qing Guo$^{1,5*\dagger}$, \ Xiaoguang Li$^{2*}$, \  Felix Juefei-Xu$^{3}$, \ Hongkai Yu$^{4}$, \ Yang Liu$^{5,6\dagger}$, \ Song Wang$^{2}$}

\affiliation{\institution{
$^{1}$ College of Intelligence and Computing, Tianjin University, China \\ 
$^{2}$ University of South Carolina, USA \ $^{3}$Alibaba Group, USA \ $^{4}$Cleveland State University, USA}}
\affiliation{\institution{$^{5}$Nanyang Technological University, Singapore \ $^{6}$ Zhejiang Sci-Tech University, China}}
\thanks{$^*$ Qing Guo and Xiaoguang Li are co-first authors and contribute equally. $\dagger$ Qing Guo and Yang Liu are the corresponding authors (\email{tsingqguo@ieee.org}{tsingqguo@ieee.org})}

\renewcommand{\shortauthors}{Qing Guo et al.}

\begin{abstract}
Image inpainting aims to restore the missing regions of corrupted images and make the recovery result identical to the originally complete image, which is different from the common generative task emphasizing the naturalness or realism of generated images.
Nevertheless, existing works usually regard it as a pure generation problem and employ cutting-edge deep generative techniques to address it. The generative networks can fill the main missing parts with realistic contents but usually distort the local structures or introduce obvious artifacts. 
In this paper, for the first time, we formulate image inpainting as a mix of two problems, \ie, predictive filtering and deep generation.
Predictive filtering is good at preserving local structures and removing artifacts but falls short to complete the large missing regions.
The deep generative network can fill the numerous missing pixels based on the understanding of the whole scene but hardly restores the details identical to the original ones.
To make use of their respective advantages, we propose the \textit{joint predictive filtering and generative network (JPGNet)} that contains three branches: \textit{predictive filtering~\&~uncertainty network (PFUNet)}, \textit{deep generative network}, and \textit{uncertainty-aware fusion network (UAFNet)}.
The PFUNet can adaptively predict pixel-wise kernels for filtering-based inpainting according to the input image and output an uncertainty map. This map indicates the pixels should be processed by filtering or generative networks, which is further fed to the UAFNet for a smart combination between filtering and generative results.
Note that, our method as a novel framework for the image inpainting problem can benefit any existing generation-based methods. 
We validate our method on three public datasets, \ie, Dunhuang, Places2, and CelebA, and demonstrate that our method can enhance three state-of-the-art generative methods (\ie, StructFlow, EdgeConnect, and RFRNet) significantly with slightly extra time costs. We have released the code at \href{https://github.com/tsingqguo/jpgnet}{https://github.com/tsingqguo/jpgnet}.
\end{abstract}

\ccsdesc[500]{Computing methodologies~Computer vision}

\keywords{Image Inpainting, Predictive Filtering, Generative Network}


\begin{teaserfigure}
\centering
  \includegraphics[width=0.98\linewidth]{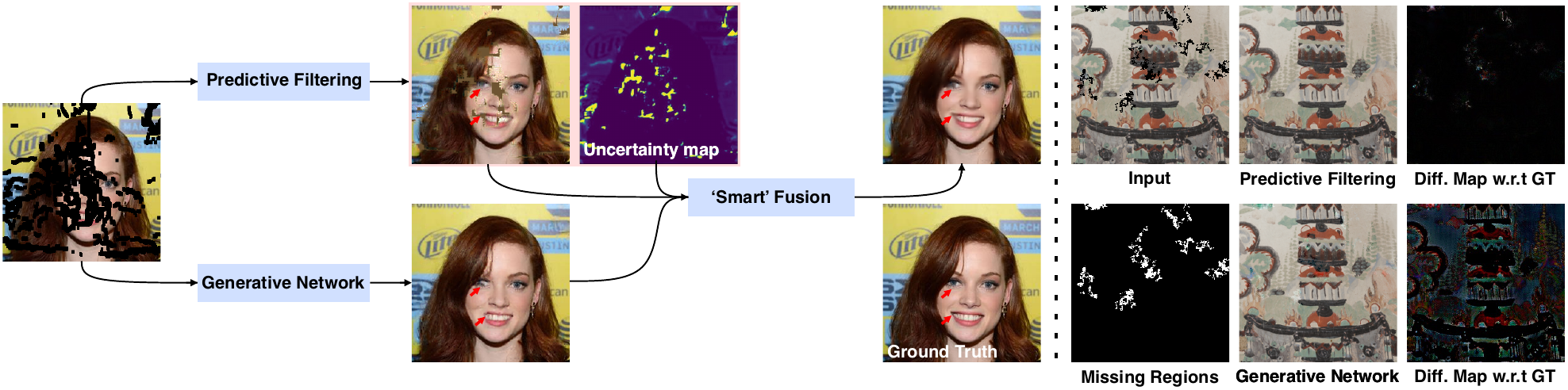}
  \vspace{-1em}
  \caption{Left: overview of the proposed inpainting framework. Predictive filtering is able to recover the local structures (\eg, the girl's eye and mouth) while the generative network (\eg, RFRNet \cite{li2020recurrenta}) is good at filling large missing regions according to the understanding of the whole face. Our method takes their respective advantages, preserving the local structure and filling all missing parts naturally. Right: An example of completing the mural image from Mogao Grottoes, Dunhuang \cite{yu2019dunhuang}, China via the predictive filtering and another generative network (\ie, StructFlow \cite{ren2019structureflow}). Although the results of both methods look the same with the ground truth, the generative network may change the intensity of unmissed pixels according to the difference map between the prediction results and ground truth. In contrast, predictive filtering only makes changes around the missing regions. Note that, under the mural restoration task, the fidelity to the ground truth is significantly important.}
  \label{fig:teaser}
\end{teaserfigure}

\maketitle

\section{Introduction}\label{sec:intro}

Image inpainting is to restore the missing or damaged areas in an image \cite{bertalmio2000image,guillemot2013image}, and it has been widely studied in computer vision and computer graphics communities long before this wave of deep generative modeling of the images. There have been numerous applications involving image inpainting such as restoring images from text overlays \cite{modha2014image}, occlusion removal for biometrics applications \cite{btas16_fastfood,cvprw14_hallucinate,fx_dissertation,accv18_rankgan,pr19_ssr2}, object removal or replacement \cite{criminisi2004region}, high-realisticity photo editing \cite{richard2001fast} using \eg, Photoshop, \etc. Image inpainting itself is a hard problem due to fact that it is ill-posed and does not have a unique solution for any particular inpainting problem, \ie, the infamous one-to-many setting. To this end, priors and/or image-specific assumptions need to be introduced to regulate the solutions.

%
\begin{figure*}[t]
\centering
\includegraphics[width=1.0\linewidth]{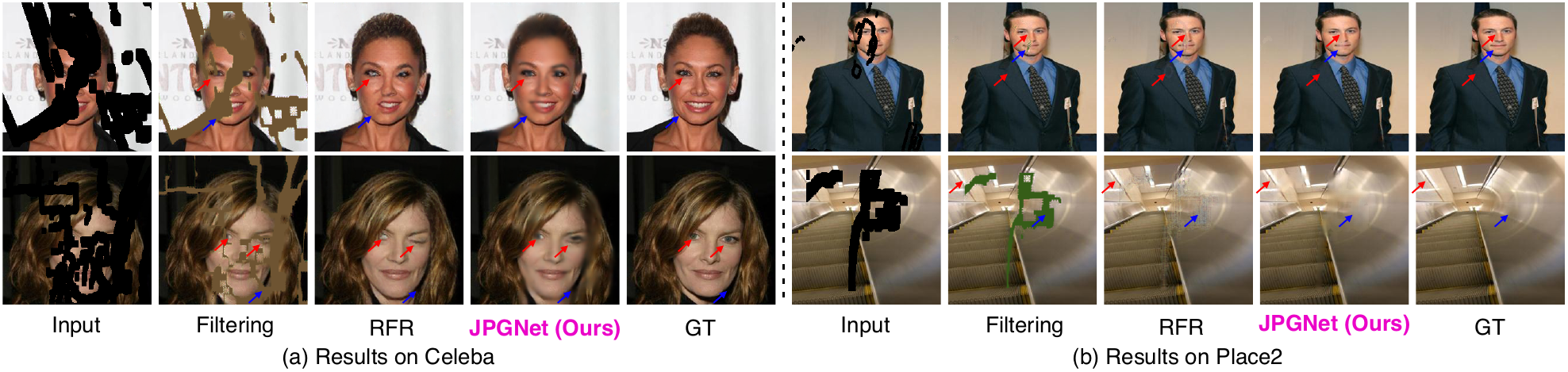}
\vspace{-2em}
\caption{Visualization results of \textit{predictive filtering}, deep generative network, and \textit{joint predictive filtering and generative network (JPGNet)} on the CelebA and Place2 datasets. We set the deep generative network as the state-of-the-art method, \ie, RFRNet \cite{li2020recurrenta}. The red arrows show the advantages of predictive filtering for handling small missing regions and recovering local structures. The blue arrows present that generative network is good at filling large missing regions. The final results show that the proposed JPGNet share both advantages without the texture artifacts in RFRNet. }\label{fig:observations}
\vspace{-1em}
\end{figure*}
%

In recent years, deep generative image modeling approaches such as generative adversarial networks (GAN) variants \cite{yu2018generative,accv18_rankgan} as well as variational autoencoder (VEA) variants \cite{yeh2017semantic,peng2021generating}, autoregressive variants \cite{oord2016conditional,salimans2017pixelcnn++}, \etc, have shown strong performance on the image inpainting problem. These deep generative approaches can learn from a large-scale image corpus and produce novel images that share the similar image distribution as the ones in the training dataset. This strong capability of generating novel images, when conditioned upon some visible regions in the image, naturally gives solutions to the image inpainting problems.
Perhaps the most impressive advantage of deep generative approaches is that they are good at dealing with large missing area image inpainting problem because they are originally intended for generating an entire image. 
Nevertheless, image inpainting is actually an image restoration problem, rather than a pure image generation task. The latter one only cares about whether the generated image is realistic or natural while the former one requires that the recovered regions to be identical to the true ones.
Such differences make the state-of-the-art generative network-based methods less perfect for image inpainting task since they may inpaint realistic but less faithful results. 
For example, even the state-of-the-art method, \eg, RFRNet \cite{li2020recurrenta}, can produce distorted local structures (See \figref{fig:teaser}). 
Moreover, some real-world applications, \eg, mural restoration \cite{yu2019dunhuang}, mainly focus on the recovery fidelity instead of the naturalness. Nevertheless, the generative network-based method (\eg, StructFlow \cite{ren2019structureflow}) modifies the unmissed pixels significantly (See Right subfigure in \figref{fig:teaser}) although it also leads to naturally-looking inpainting result.

On a parallel track, deep predictive filtering approaches have showcased its effectiveness on various restoration tasks such as denoising \cite{BakoACMTOG2017,MildenhallCVPR2018,guo2013image}, deraining \cite{guo2020efficientderain}, and shadow removal \cite{fu2021auto}. The key idea is to generate pixel-wise kernels via a deep CNN, which are then used to guide the reconstruction of each pixel with the support of its neighboring pixels.
The main advantage of predictive filtering is that it respects the main information of the input, meeting the high fidelity requirements of the image inpainting. Nevertheless, it falls short to generate missing pixels from nothing or where there is not sufficient nearby pixels to extract information from.
To this date, the use of deep predictive filtering for the task of image inpainting has not been explored yet. 


In this paper, for the first time, we take the best of both worlds and formulate image inpainting as a mix of two problems, \ie, predictive filtering and deep generation. Predictive filtering is good at preserving local structures and removing artifacts but falls short to complete the large missing regions. The deep generative network, as discussed above, can fill the numerous missing pixels based on the understanding of the whole scene but hardly restores the details identical to the original ones.
To make use of their respective advantages, we propose the \textit{joint predictive filtering and generative network (JPGNet)} that contains three branches: \textit{predictive filtering~\&~uncertainty network (PFUNet)}, \textit{deep generative network}, and \textit{uncertainty-aware fusion network (UAFNet)}.
The PFUNet can adaptively predict pixel-wise kernels for filtering-based inpainting according to the input image and output an uncertainty map. This map indicates the pixels should be processed by filtering or generative networks, which is further fed to the UAFNet for a smart combination between filtering and generative results.
Note that, our method as a novel framework for the image inpainting problem can benefit any existing generation-based methods. 
We validate our method on three public datasets, \ie, Dunhuang, Places2, and CelebA, and demonstrate that our method can enhance three state-of-the-art generative methods (\ie, StructFlow, EdgeConnect, and RFRNet) significantly with slightly extra time cost.

\section{Related Work}\label{sec:related}

\subsection{Generative Network for Image Inpainting}

In recent years, deep neural network (DNN)-based approaches have displayed a promising performance in the image inpainting field \cite{elharrouss2020imagea}. The early work \cite{pathak2016context} trained a CNN for completing large holes though context encoders. Then, the method \cite{iizuka2017globally} introduced the globally and locally consistent for image completion through both global and local discriminators as adversarial losses. The global discriminator assesses whether the completed image is coherent as a whole, while the local discriminator focuses on a small area centered at the generated region to enforce the local consistency.        
Meanwhile, there have been several studies focusing on generative face inpainting. Specifically, Yu \etal\cite{yu2018generative}  found that convolutional neural networks are less effective in building long-term correlations for image inpainting. To solve this problem, they proposed contextual attention to borrow features from remote regions. Liu \etal\cite{liu2018image} observed that the substituting pixels in the masked holes of the inputs might introduce artifacts to the final results, and they proposed the partial convolution to force the network to use valid pixels only.
Nazeri \etal\cite{nazeri2019edgeconnect} designed a two-stage adversarial model comprising an edge generator followed by an image completion network. The edge generator hallucinates edges of the missing region of the image, and the image completion network fills in the missing regions using hallucinated edges as a prior.
Ren \cite{ren2019structureflow} also split the inpainting task into two stages, \ie, structure re-construction and texture generation. The first stage completes the missing structures of the inputs. In the second stage, a texture generator yields image details based on the reconstructed structures. 
More recently, Li \cite{li2020recurrenta} proposed to recurrently infer the hole boundaries of the convolutional feature maps and use them as clues for further inference. The module progressively strengthens the constraints for the hole center and the results become explicit.

Although presenting impressive inpainting results, above works usually regard the image inpainting as a pure generation problem and employ cutting-edge deep generative techniques to address it, which can fill the main missing parts with realistic contents but might fail to get a high identity to the original one. As shown in \figref{fig:teaser}, the state-of-the-art deep generative networks, \ie, RFRNet \cite{li2020recurrenta} and StructFlow \cite{ren2019structureflow}, might change the local structures or the pixel intensities, making the inpainting images showing obvious difference w.r.t. to ground truth. In this work, for the first time, we formulate the image inpaint as a mix of two tasks, \ie, predictive filtering and deep generation. By taking their respective advantages, our method can not only recover the local structures but also fill all missing parts naturally.

\subsection{Background-aware Image Inpainting} 

A typical idea for image inpainting is to find suitable patches in unmissed regions to fill the hole. The key problem is how to get related patches and how to complete missing regions in a natural way \cite{barnes2009patchmatch}. Ružić and Pižurica \cite{ruzic2015context} searched the well-matched patch via Markov random filed. Jin \etal\cite{jin2015annihilating} and Guo \etal\cite{guo2017patch} utilized the low rank theory for patch searching and filling. Kawai \etal \cite{kawai2015diminished} selected patches according to the object and search around background for inpainting. 
Liu \etal \cite{liu2018structure} utilized the structure constraint via homograph transformation for effective inpainting.
Xue \etal \cite{xue2017depth} studied the depth-based inpainting problem via the low rank. 
Ding \etal\cite{ding2019image} proposed to fill missing regions via the non-local texture matching and nonlinear filtering. 
In this paper, we also employ the filtering for inpainting that utilizes the missing pixels' neighboring pixels for reconstruction. In contrast to \cite{ding2019image}, our filters are pixel-wise and estimated in a one-step way via a pre-trained encoder-decoder, which is able to adapt to different inputs and regions efficiently. To best of our knowledge, this is the first attempt to adapt the deep predictive filtering for image inpainting.

\subsection{Predictive Filtering for Image Restoration}
Deep predictive filtering is a recently proposed DNN framework for image processing \cite{BakoACMTOG2017,mildenhall2018burst,MildenhallCVPR2018,guo2020efficientderain,brooks2019learning,guo2020watch,fu2021auto,cheng2020adversarial,zhai2020s}. It is basic idea is to predict the kernels for all pixels via a pre-trained deep network and use the kernels to conduce pixel-wise image filtering. Bako \etal \cite{BakoACMTOG2017}  and Mildenhall \etal \cite{mildenhall2018burst} used the predictive filtering for denoising. Guo \etal\cite{guo2020efficientderain} extended it to the field of deraining while Fu \etal\cite{fu2021auto} regarded it as a tool for fusing multiple exposure images to realize efficient shadow removal. In addition to image restoration problem, Brooks \etal\cite{brooks2019learning} and Guo \etal \cite{guo2020watch} employ the predictive filtering for synthesizing the motion blur. 
In contrast to the degradation like noise, rain streaks, and shadow, the holes (\ie, missing regions) make the image information totally lost. Intuitively, the small holes or the missing pixels at the boundary of a big hole can be reconstructed by their neighboring pixels. Nevertheless, the pixels at the center of the large hole could have diverse possibility and cannot be recovered from the neighboring pixels. These pixels should be `guessed' by understanding the whole scene. Hence, it is difficult to achieve high performance for image inpainting via the naive implementation of predictive filtering. In this work, in contrast to existing predictive filtering-based restoration methods, we propose to combine it with the generative network via a new design fusion network.

\section{Intuitive Discussion}\label{sec:prelim}
Given an image $\mathbf{I}\in\mathds{R}^{H\times W}$ whose partial pixels are missed, we aim to fill these regions with realistic contents and make the processed image identical to the originally completed counterpart (\ie, the ground truth of image inpainting) denoted as $\mathbf{I}^*\in\mathds{R}^{H\times W}$.
In this section, we first introduce two solutions for image inpainting, \ie, \textit{predictive filtering and generative network}. Then, we provide some intuitive comparisons between these two solutions, driving the motivation of this work.

\subsection{Predictive Filtering for Image Inpainting} \label{subsec:predfilter}
\textit{Predictive filtering} is an advanced image filtering technology that combines the advantages of the classical image filtering and deep learning and has been validated on various tasks, \eg, image denoising \cite{BakoACMTOG2017,MildenhallCVPR2018}, deraining \cite{guo2020efficientderain}, shadow-removal \cite{fu2021auto}, and blurring \cite{guo2020watch}. 
Its basic pipeline is to filter image with pixel-wise kernels predicted from a deep convolutional neural network (CNN). 
Note that, to best of our knowledge, this is the very first attempt to formulate the image inpainting as a predictive filtering problem.

%
\begin{figure}[t]
\centering
\includegraphics[width=0.9\linewidth]{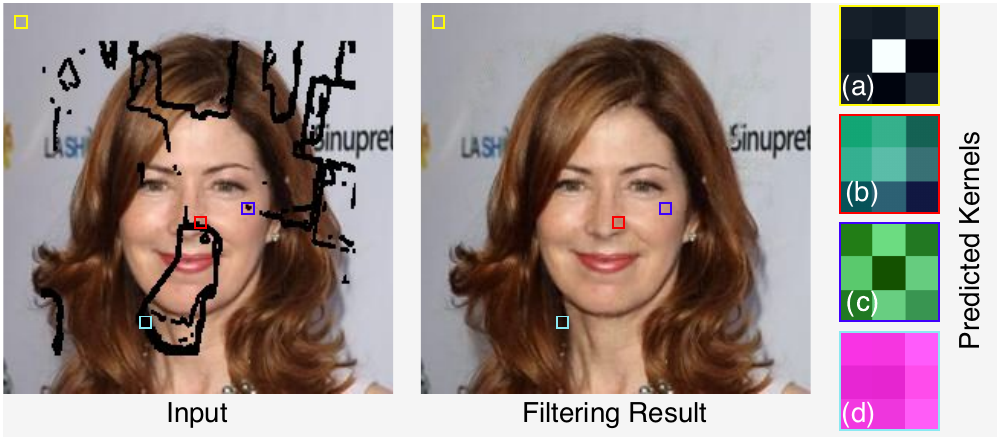}
\vspace{-1em}
\caption{ Predicted kernels for different pixels. Clearly, the kernels' patterns are related to the missing regions, directly.}\label{fig:kernels}
\vspace{-1em}
\end{figure}
%

Specifically, given the input corrupted image $\mathbf{I}$, we process it with pixel-wise kernels by
%
\begin{align}\label{eq:pixelfilter}
\hat{\mathbf{I}}_{1}(\mathbf{p}) = \sum_{\mathbf{q}\in\mathcal{N}_\mathbf{p}}\mathbf{K}_{p}(\mathbf{q}-\mathbf{p})\mathbf{I}(\mathbf{q}),
\end{align}
%
where $\mathbf{p}$ and $\mathbf{q}$ represent the pixel coordinates, $\mathcal{N}_\mathbf{p}$ denotes the set of neighboring pixels of $\mathbf{p}$, and
$\mathbf{K}_p\in\mathds{R}^{K\times K}$ is the exclusive kernel for pixel $\mathbf{p}$ with its kernel size being $K$. 
Intuitively, Eq.~\eqref{eq:pixelfilter} is to reconstruct the pixel $\mathbf{p}$ with its neighboring pixels in $\mathcal{N}_\mathbf{p}$ and the kernel $\mathbf{K}_{p}(\mathbf{q}-\mathbf{p})$ determines the combination weight on the pixel $\mathbf{q}$. The size of $\mathcal{N}_\mathbf{p}$ is $K^2$.
Different from the traditional image filtering methods (\eg, Sobel) using fixed kernels for all pixels, the predictive filtering employs a deep CNN to adaptively predict the pixel-wise kernels according to the input: 
%
\begin{align}\label{eq:kpn}
\mathbf{K} = \text{UNet1}(\mathbf{I}),
\end{align}
%
where $\mathbf{K}\in \mathds{R}^{H\times W \times K^2}$ contains the kernels of all pixels, \ie,  $\{\mathbf{K}_p\}$ while $\text{UNet1}(\cdot)$ is an UNet architecture \cite{guo2020efficientderain}.
We can train the network with image quality related loss functions, \eg, $L_1$ and SSIM loss functions.
As the results shown in the \figref{fig:observations}, the predictive filtering method is able to fill the small regions accurately but fails to predict the large missing patches.
This is reasonable since the filtering is to reconstruct the missing pixel according to its neighbors while the large missing patches should be filled through the capability of understanding the whole scene.

\subsection{Generative Network for Image Inpainting}\label{subsec:generativenetwork}
With the rapid development of deep generative networks, state-of-the-art methods design novel generative networks \cite{nazeri2019edgeconnect,ren2019structureflow} to map the input degraded images to the completed counterparts directly. They usually train their models with the adversarial loss function. Such methods are able to reason the large missing regions since the pre-trained deep generative models can deeply understand the scene in the degraded image and inject the prior information into the miss regions. 
Nevertheless, these generative models are less effective on filling small regions, which may generate new structures that are not identical to the completed counterpart.
As shown in the first case of \figref{fig:observations}, the generative method leads to distorted faces, making inpainting results not like the ground truth.

\subsection{Observations and Motivations}
%

{\bf Observations.} We use the predictive filtering in \secref{subsec:predfilter} and a state-of-the-art deep generative network, \ie, RFRNet \cite{li2020recurrenta}, to conduct the inpainting, respectively. We show visualization results on the CelebA and Place2 datasets in \figref{fig:observations} and observe that: \ding{182} the predictive filtering is able to recover the small missing structures effectively while the generative network is ease to distort them or introduces artifacts. As shown in the results on CelebA, the predictive filtering can recover the missing eyes that have similar structure (See the opening status and eyeball position) with the ground truth. In contrast, the generative network makes the opening eyes become close and normal eyes be distorted. We have similar observations on the Place2. \ding{183} The deep generative network is able to fill all missing pixels while the filtering fails on the large missing regions. \ding{184} The RFRNet has obvious artifacts on the generated regions while the filtering results look clear on restored regions. 

%
\begin{figure}[t]
\centering
\includegraphics[width=0.9\linewidth]{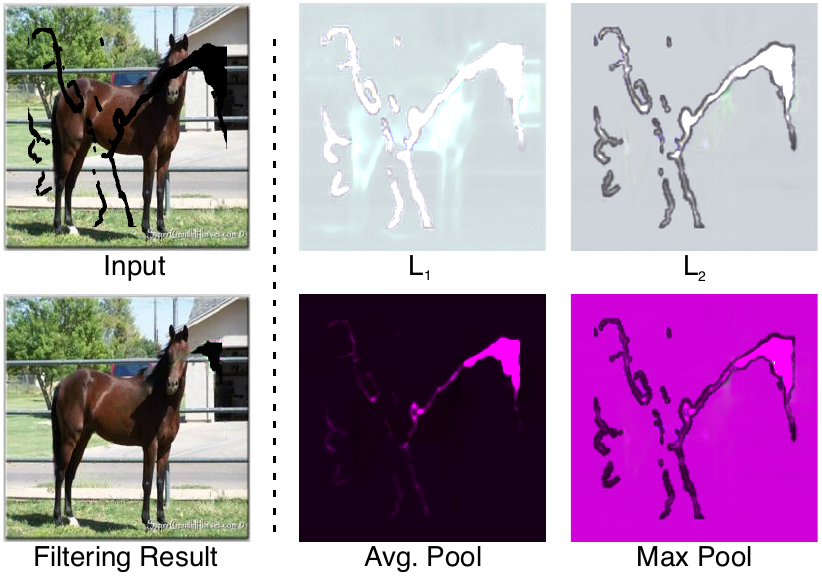}
\vspace{-1em}
\caption{Uncertainty maps based on average pool, max pool, $L_1$,  and $L_2$. We visualize the input and filtering results, showing that the map based on average pool can indicate whether the pixels could be recovered via the filtering.}\label{fig:uncerntainty}
\vspace{-1em}
\end{figure}
%

%
\begin{SCfigure*}
\centering
\includegraphics[width=0.79\textwidth]{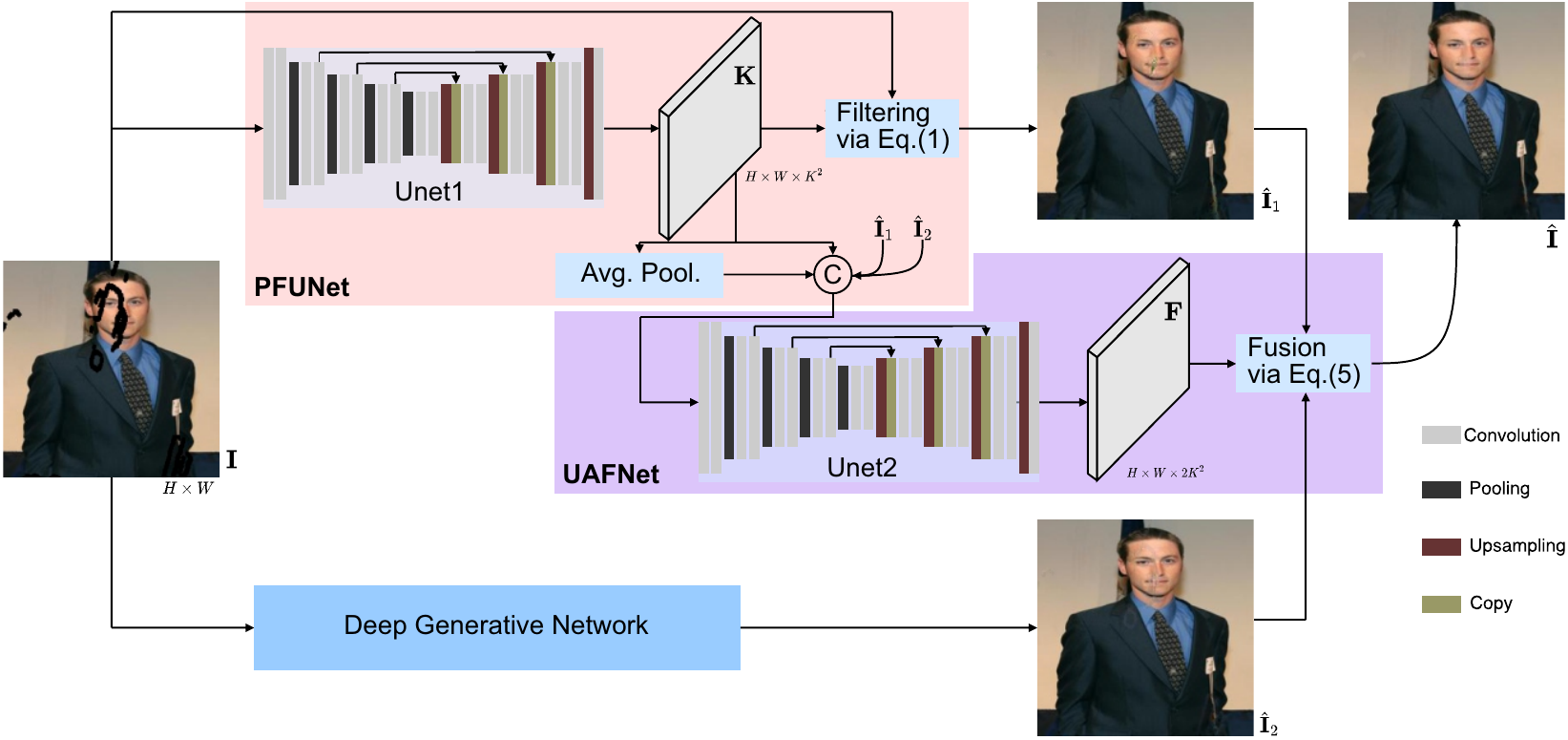}
\caption{Overview of the joint predictive filtering and generative network (JPGNet) containing three modules, \ie, predictive filtering \& uncertainty network (PFUNet), deep generative network, and uncertainty-aware fusion network. The `\textcircled{c}' denotes the concatenation layer. The `Avg. Pool.' represents the averaging pooling along the channel dimension of $\mathbf{K}$, outputting the uncertainty map.}
\label{fig:pipeline}
\end{SCfigure*}
%

{\bf Motivations and Challenges.} Clearly, above two solutions have their respective advantages and disadvantages. To let the inpainting be able to restore the small structures while reasoning large missing regions, it is better to make a combination and solve two major issues: \ding{182} how to identify the regions that can be reconstructed by filtering or generative networks? \ding{183} how to fuse the results of filtering and generation in a `smart' way ? 
For the first challenge, we propose the \textit{predictive filtering~\&~uncertainty network} by extending the predictive filtering without extra costs. It empowers the naive predictive filtering to estimate an uncertainty map which assigns each pixel a score indicating whether the pixel can be or need to be reconstructed by filtering. 
For the second challenge, we propose the \textit{uncertainty-aware fusion network} that can fuse the results of predictive filtering and generative network.
We will detail the whole framework and contributions in \secref{sec:method}.

\section{Joint Predictive Filtering and \\ Generative Network}\label{sec:method}

\subsection{Overview} 
The proposed method contains three modules, \ie, \textit{predictive filtering~\&~uncertainty network (PFUNet)}, \textit{deep generative network (DGNet)},  and \textit{uncertainty-aware fusion network (UAFNet)}. 
Intuitively, given a corrupted image $\mathbf{I}$, we feed it to the PFUNet and \textit{deep generative network}, respectively, and we get two initial inpainting results, \ie, $\hat{\mathbf{I}}_1$ and $\hat{\mathbf{I}}_2$. 
Note that, the PFUNet also outputs an uncertainty map $\mathbf{U}$ according to the predicted filters, which assigns each pixel a score indicating whether the pixel can be or need to be reconstructed by filtering.
Then, we concatenate the uncertainty map, predicted kernels, and initial restored images and feed them into the UAFNet to guide the fusion of $\hat{\mathbf{I}}_1$ and $\hat{\mathbf{I}}_2$, leading to the final result $\hat{\mathbf{I}}$ .
Please refer to \figref{fig:pipeline} for understanding the whole pipeline easily.
Note that, our method can help any generative inpainting methods by setting them as the \textit{deep generative network} in our framework. 
In the following, we detail the \textit{PFUNet} and \textit{UAFNet}, respectively.

%
\begin{figure}[t]
\centering
\includegraphics[width=1.0\linewidth]{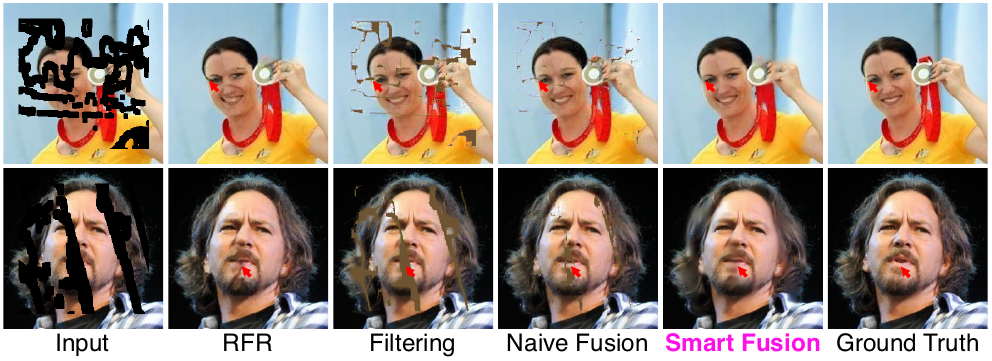}
\vspace{-1em}
\caption{Two examples from CelebA dataset with the inpainting results of the deep generative network (\eg, RFRNet), predictive filtering, naive fusion via \reqref{eq:naivefusion}, and smart fusion via UAFNet in \secref{subsec:uncerntainty}.}\label{fig:fusioncmp}
\vspace{-1em}
\end{figure}
%

\subsection{Predictive filtering~\&~uncertainty network} \label{subsec:predict_uncertainty}
In this section, we first rethink the predictive filtering in \secref{sec:prelim} to understand the functionality of predicted kernels, inspiring how to construct an uncertainty map indicating whether a pixel could be reconstructed by neighboring pixels.

{\bf Rethinking the functionality of pixel-wise kernels.}
Considering the filtering function in \reqref{eq:pixelfilter}, we have the following inferences: \ding{182} when the pixel $\mathbf{p}$ in the input image is not missed, its kernel (\ie, $\mathbf{K}_p$) should be specified as
%
\begin{align}\label{eq:identifykernel}
\mathbf{K}_p(\mathbf{q}-\mathbf{p})= \left\{\begin{matrix}
 1,~\text{if}~\mathbf{q}-\mathbf{p}=[\frac{K-1}{2},\frac{K-1}{2}],\\ 
0,\quad \quad \quad \quad \quad \text{otherwise}.
\end{matrix}
\right.
\end{align}

%
to make sure the $\hat{\mathbf{I}}(\mathbf{p})=\mathbf{I}(\mathbf{p})$ (See the kernel (a) in \figref{fig:kernels}). \ding{183} when the pixel $\mathbf{p}$ at the boundary of the missing regions, its kernel is definitely different from the one in \reqref{eq:identifykernel} and is related to the shape of the regions (See the (b) and (c) kernels in \figref{fig:kernels}).  \ding{184} Moreover, the kernel of the pixels near to the center of the large missing regions present uniform values, having different patterns to those of other pixels (See the kernel (d) in  \figref{fig:kernels}.). 
According to above discussion, the patterns of predicted kernels are able to represent the capability of using filtering to reconstruct the missing pixels. 

{\bf Uncertainty map prediction and naive fusion.} To validate above hypothesis, we consider operations including $\{\text{max pool}, \\ \text{average pool}, L_1, L_2\}$ to process each kernel and get a score for each pixel to indicate whether this pixel could be recovered via the filtering or not.
As a result, we can obtain a map that is further normalized to [0,1] and denoted as the uncertainty map $\mathbf{U}$.
We show an example in \figref{fig:uncerntainty} with four maps based on max pool, average pool, $L_1$, and $L_2$, respectively.
Comparing the four maps with the filtering result, the map based on the average pool is able to represent the capability of pixel-wise filtering to reconstruct the missing pixels effectively. 
For example, the highlighted regions in `Avg. Pool' is identical to the non-recovered or under-recovered regions in the filtering results.

Intuitively, we can use the uncertainty map to fuse the filtering and generation results (\ie, $\hat{\mathbf{I}}_1$ and $\hat{\mathbf{I}}_2$) directly, \ie,
%
\begin{align}\label{eq:naivefusion}
\hat{\mathbf{I}} = (1-\mathbf{U})\odot\hat{\mathbf{I}}_1 + \mathbf{U}\odot\hat{\mathbf{I}}_2,
\end{align}
%
where `$\odot$' denotes the element-wise multiplication. The above fusion process is to select the results of generative network or filtering as the final counterparts according to the uncertainty map. The pixels with high uncertainty values means they cannot be recovered via the filtering and rely on  the deep generative network. In contrast, the pixels with low uncertainty are determined by filtering results.
As the results shown in \figref{fig:fusioncmp}, the naive fusion is able to limit the distortion introduced by the deep generative network as the regions indicated by the red arrows.
Nevertheless, such fusion process ignores the potential relationship across neighboring pixels, leading to a lot of remaining uncompleted regions. 
In the following, we train a uncertainty-aware fusion network to fuse the filtering and generation results in a `smart' way.

\subsection{Uncertainty-aware fusion network}  \label{subsec:uncerntainty}
Given two initial results, \ie, $\hat{\mathbf{I}}_1$ and $\hat{\mathbf{I}}_2$, we propose to fuse them by
%
\begin{align}\label{eq:advancedfusion}
\hat{\mathbf{I}}(\mathbf{p}) = \sum_{\mathbf{q}\in\mathcal{N}_\mathbf{p}}\mathbf{F}_{p}(\mathbf{q}-\mathbf{p})\odot[\hat{\mathbf{I}}_1(\mathbf{q}), \hat{\mathbf{I}}_2(\mathbf{q})],
\end{align}
%
where $\mathbf{p}$ and $\mathbf{q}$ are the pixel coordinates, $\mathcal{N}_\mathbf{p}$ denotes the set of neighboring pixels of $\mathbf{p}$. The matrix $\mathbf{F}_p\in \mathds{R}^{K\times K\times 2}$ contains the fusion parameters with $K^2$ being the size of $\mathcal{N}_\mathbf{p}$ and $\mathbf{F}_p(\mathbf{q}-\mathbf{q})$ and $[\hat{\mathbf{I}}_1(\mathbf{q}), \hat{\mathbf{I}}_2(\mathbf{p})]$ are vectors with size of $1\times 2$. Comparing with the \reqref{eq:naivefusion}, the new fusion formulation consider the neighboring pixels. Moreover, we propose to predict the parameters $\mathbf{F}_p$ via a network that takes the uncertainty map $\mathbf{U}$, predicted kernels $\mathbf{K}$, and the initial results (\ie, $\hat{\mathbf{I}}_1$ and $\hat{\mathbf{I}}_2$) as the inputs and outputs
%
\begin{align}\label{eq:fusionnetwork}
\mathbf{F} = \text{UNet2}(\mathbf{U}, \mathbf{K}, \hat{\mathbf{I}}_1, \hat{\mathbf{I}}_2),
\end{align}
%
$\mathbf{F}\in \mathds{R}^{H\times W\times 2K^2}$. It contains all fusion parameters, \ie, $\{\mathbf{F}_p\}$. We will detail the network architecture in \secref{subsec:impldetails}.

\subsection{Implementation Details} \label{subsec:impldetails}
We implement our method to enhance three state-of-the-art deep generative networks,  \ie, StructFlow \cite{ren2019structureflow}, EdgeConnect \cite{nazeri2019edgeconnect}, and RFRNet \cite{li2020recurrenta}, we follow their released code and train their models according respective default setups. In the following, we introduce the details about the proposed PFUNet and UAFNet. 

{\bf Network Architectures.} In PFUNet and UAFNet, the UNet1 and UNet2 are their main architectures and we follow the UNet \cite{ronneberger2015u} setups in \cite{guo2020efficientderain}. They share similar structures but have different input and output channels according to their functionalities. Please refer to \tableref{tab:arch} for details. Note that, in previous subsections, we formulate the method for single-channel images for better understanding. Here, we implement our method for the color images by predicting kernels or fusion parameters for each color channel independently. For example, the size of $\mathbf{K}$ and $\mathbf{F}$ becomes $3K^2$ and $2*3K^2$. We fix the kernel size $K$ as 3 since it allows efficient training.

%
\begin{table}[t]
\footnotesize
\centering
\caption{Architecture of UNet1 or UNet2 for PFUNet or UAFNet, respectively. The `Blk' denotes a block containing three stacked convolution layers. The `[$\cdot$]' denotes the concatenation operation. Each convolution layer is followed by a batch normalization layer and an activation layer. For UNet1, the input is the corrupted image $\mathbf{I}$ with $C_\text{in}=3$ for color image and the output is $\mathbf{K}$ with $C_\text{o}=3K^2=27$ for each pixel in each color channel has a $K\times K$ kernel. For UNet2, the input is [$\mathbf{U}$, $\mathbf{K}$, $\hat{\mathbf{I}}_1$, $\hat{\mathbf{I}}_2$] and the output is $\mathbf{F}$ with $C_\text{o}=2*3K^2$.}\label{tab:arch}
\begin{tabular}{l|c|c|l|l}
\toprule
 & input & output & output size  & 49-layer \\\midrule
Blk1  & Input & $\mathbf{x}_1$   & $256\times256$      & $3\times$~conv{(}3, $C_\text{in}$, 64{)}              \\
Blk2  & $\mathbf{x}_1$ & $\mathbf{x}_2$ & $128\times128$      & $3\times$~conv{(}3, 64, 128{)}             \\
Blk3  & $\mathbf{x}_2$ & $\mathbf{x}_3$ & $64\times64$        & $3\times$~conv{(}3, 128, 256{)}           \\
Blk4  & $\mathbf{x}_3$ & $\mathbf{x}_4$ & $32\times32$        & $3\times$~conv{(}3, 256, 512{)}           \\
Blk5  & $\mathbf{x}_4$ & $\mathbf{x}_5$ & $16\times16$        & $3\times$~conv{(}3, 512, 512{)}           \\
Blk6  & $\mathbf{x}_5$ & $\mathbf{x}_6$ & $32\times32$        & $3\times$~conv{(}3, 512, 512{)}           \\
Blk7  & [$\mathbf{x}_6$,$\mathbf{x}_4$] & $\mathbf{x}_7$& $64\times64$  & $3\times$~conv{(}3, 512, 256{)} \\
Blk8  & [$\mathbf{x}_7$,$\mathbf{x}_3$] & $\mathbf{x}_8$ & $128\times128$ & $3\times$~conv{(}3, 256, 27{)} \\
Conv    & [$\mathbf{x}_8$,$\mathbf{x}_2$] & $\mathbf{K}$/ $\mathbf{F}$ & $256\times256$ & conv{(}1, 27, $C_\text{o}${)} \\
\bottomrule
\end{tabular}
\end{table}
%

%
\begin{figure*}[t]
\centering
\includegraphics[width=0.9\linewidth]{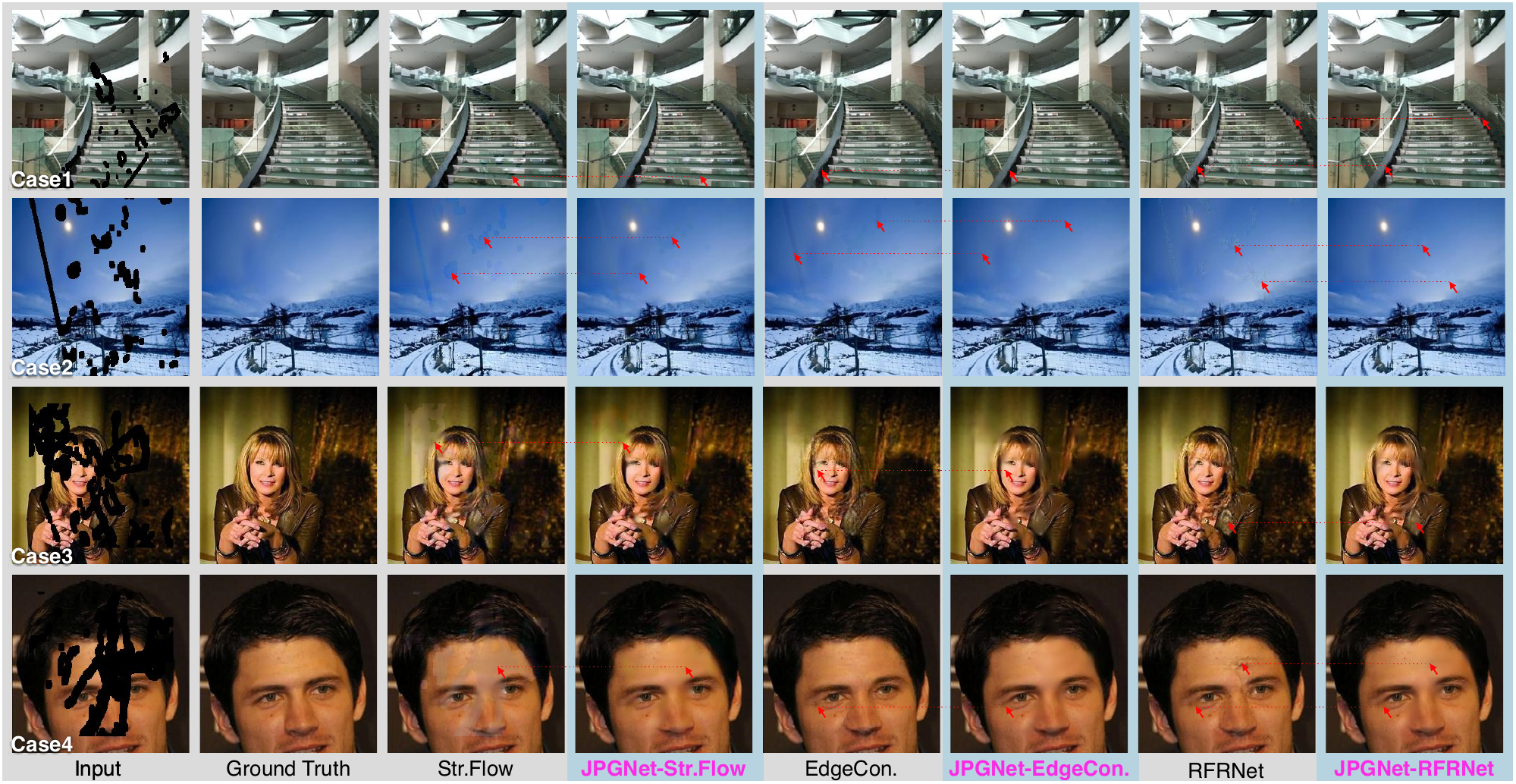}
\caption{Four visualization results of StructFlow (Str.~Flow) \cite{ren2019structureflow}, EdgeConnect (EdgeCon.) \cite{nazeri2019edgeconnect}, RFRNet \cite{li2020recurrenta}, and the JPGNet-enhanced counterparts. The first two cases are from the Place2 dataset while Case3 and Case4 stem from the CelebA dataset. We use red arrows to show the main advantages of the JPGNet.}\label{fig:viscmp}
\end{figure*}
%

{\bf Loss functions.} We train the whole framework with two stages. First, we train the PFUNet by considering two loss functions, \ie, $L_1$ and SSIM loss functions. Given the estimated image $\hat{\mathbf{I}}$ and the ground truth $\mathbf{I}^*$, we have 
%
\begin{align}\label{eq:loss}
\mathcal{L}(\hat{\mathbf{I}}, \mathbf{I}^*) = \|\hat{\mathbf{I}}-\mathbf{I}^*\|_1-\lambda~\text{SSIM}(\hat{\mathbf{I}}, \mathbf{I}^*),
\end{align}
%
where we fix $\lambda=0.2$ during training process. After training the PFUNet and the deep generative networks, we fix them and train the UAFNet through the same loss function.

{\bf Training details.} We employ Adam as the optimizer with the learning rate of 0.0002. We train PFUNet for 450,000 epochs, and UAFNet for 100,000 epochs. All experiments are implemented on the same platform with a single NVIDIA Tesla V100 GPU.

\section{Experimental Results}\label{sec:experiments}
%
\begin{figure}[t]
\centering
\includegraphics[width=0.95\linewidth]{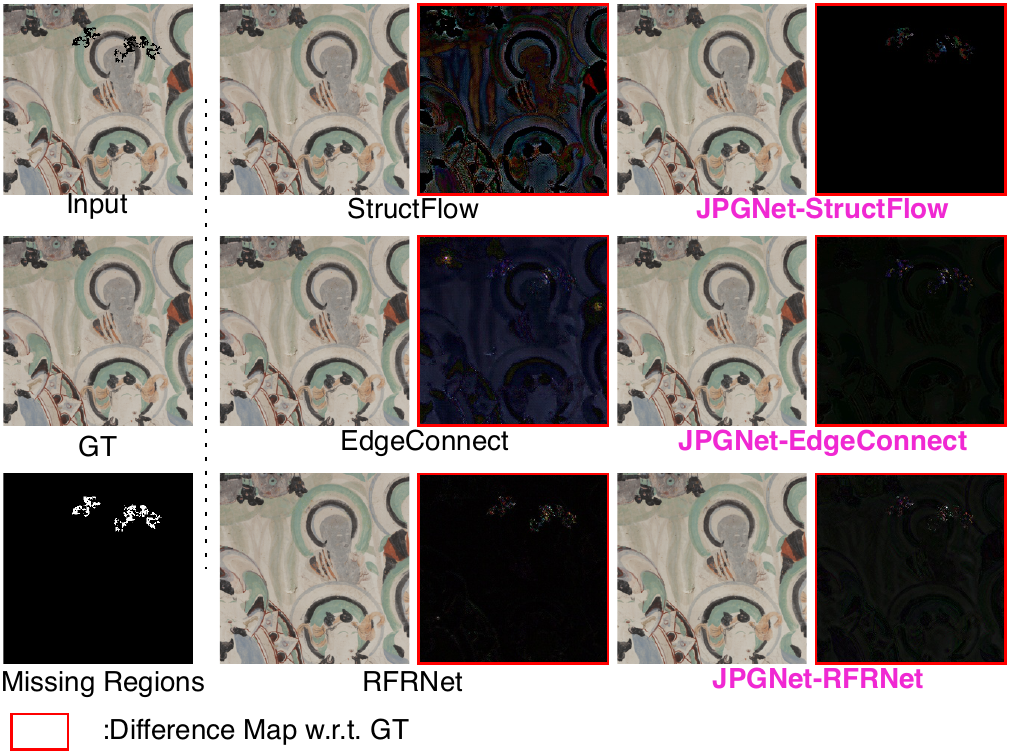}
\vspace{-1em}
\caption{An example from the Dunhuang dataset. In addition to the inpainting results of six methods, we show the difference maps between the predicted results and the ground truth (GT). }\label{fig:dunhuang}
\vspace{-1em}
\end{figure}
%
\subsection{Setups}

{\bf Datasets.} We validate the proposed method on three datasets, \ie, Places2 challenge dataset \cite{zhou2017places}, CelebA dataset \cite{liu2015deep}, and Dunhuang Challenge \cite{yu2019dunhuang}. We choose the Dunhuang challenge since it is a representative application for image inpainting in the real world, \ie, that is, to automatically recover the missing regions in the paintings that are created by thousands and thousands of artists over 10 centuries. The Places2 dataset contains over eight million images captured under over 365 scenes, which is suitable for validating the methods on natural scenes. The CelebA dataset contains over 180 thousand face images, evaluating the capability of face inpainting.

{\bf Metrics.} We follow the comment setups in the image inpainting and use the peak signal-to-noise ratio (PSNR) and structural similarity index (SSIM) for the restoration quality evaluation. We also provide the qualitative comparison via diverse visualization.

{\bf Mask Setups.} For the Places and CelebA datasets, we use the irregular mask dataset \cite{liu2018image}, which is also a common setup in recent works \cite{ren2019structureflow}. Note that, the mask images are classified to three categories (\ie, $0\%-20\%$, $20\%-40\%$, $40\%-60\%$,) based on the hole sizes relative to the image size. For the Dunhuang dataset, we follow its official setup and use their synthetic method considering the real-world degradation caused by the corrosion and aging. 

{\bf Baselines.} We use our method to enhance three state-of-the-art inpainting methods, \ie, StructFlow \cite{ren2019structureflow}, EdgeConnectnect \cite{nazeri2019edgeconnect}, RFRNetNet \cite{li2020recurrenta}.

\subsection{Comparison Results}

\begin{SCtable*}
\centering
\caption{Comparison results on Places2, CelebA, and Dunhuang datasets.}
\small
{
    {
	\begin{tabular}{l|l|ccc|ccc|c}
		
    \toprule
     & \multicolumn{1}{c|}{Datasets} & \multicolumn{3}{c|}{Places2} & \multicolumn{3}{c|}{CelebA}  & \multicolumn{1}{c}{Dunhuang} \\
    & \multicolumn{1}{c|}{Mask Ratio} &  0\%-20\% & 20\%-40\% & 40\%-60\% & 0\%-20\% & 20\%-40\% & 40\%-60\% & Default\\
    \midrule
    \multirow{6}{*}{PSNR}  
      & StructFlow \cite{ren2019structureflow} & 29.039  & 23.111  & 19.421 & 31.618   & 25.283  & 20.829  & 35.199  \\
     &  \cellcolor{tabgray} JPGNet-StructFlow  & \cellcolor{tabgray} 31.381    & \cellcolor{tabgray}  24.889  & \cellcolor{tabgray}  20.783  & \cellcolor{tabgray}  33.781  &  \cellcolor{tabgray}  26.501  & \cellcolor{tabgray}  21.477 &  \cellcolor{tabgray} 37.268 \\
    \cline{2-9}
    & EdgeConnect \cite{nazeri2019edgeconnect} & 29.909  & 23.382  & 19.516 & 32.781 & 25.347  & 20.449  & 36.419 \\
     &  \cellcolor{tabgray}  JPGNet-EdgeConnect    & \cellcolor{tabgray}  31.337    & \cellcolor{tabgray}  24.388  & \cellcolor{tabgray} 20.283  & \cellcolor{tabgray} 34.110   & \cellcolor{tabgray} 26.351  & \cellcolor{tabgray} 21.195 & \cellcolor{tabgray} 37.369 \\
    \cline{2-9}
    & RFRNet \cite{li2020recurrenta} & 29.442  & 22.701  & 18.647 & 34.521   & 26.328  & 20.994 & 35.751\\
     & \cellcolor{tabgray} JPGNet-RFRNet  & \cellcolor{tabgray} 30.834 & \cellcolor{tabgray} 24.056  & \cellcolor{tabgray} 19.954  & \cellcolor{tabgray} 35.898   & \cellcolor{tabgray} 27.639  & \cellcolor{tabgray} 22.089 & \cellcolor{tabgray} 37.646 \\
    \midrule
    \midrule
    \multirow{6}{*}{SSIM}  & StructFlow \cite{ren2019structureflow} & 0.9335 & 0.8180  & 0.6736 & 0.9487  & 0.8598  & 0.7417  & 0.9559 \\
     & \cellcolor{tabgray}  JPGNet-StructFlow &  \cellcolor{tabgray} 0.9493    & \cellcolor{tabgray} 0.8489  & \cellcolor{tabgray} 0.7141 & \cellcolor{tabgray} 0.9640  & \cellcolor{tabgray} 0.8875  & \cellcolor{tabgray} 0.8025 & \cellcolor{tabgray} 0.9682 \\
    \cline{2-9}
    & EdgeConnect \cite{nazeri2019edgeconnect}   & 0.9398  & 0.8227  & 0.6711 & 0.9586   & 0.8689  & 0.7362 & 0.9635 \\
     & \cellcolor{tabgray} JPGNet-EdgeConnect  & \cellcolor{tabgray} 0.9516  & \cellcolor{tabgray} 0.8493  & \cellcolor{tabgray} 0.7114  & \cellcolor{tabgray} 0.9670   & \cellcolor{tabgray} 0.8909  & \cellcolor{tabgray} 0.7725  & \cellcolor{tabgray} 0.9718  \\
    \cline{2-9}
    & RFRNet \cite{li2020recurrenta} & 0.9297  & 0.7893  & 0.6156  & 0.9674   & 0.8865  & 0.7538 & 0.9616 \\
     & \cellcolor{tabgray} JPGNet-RFRNet   & \cellcolor{tabgray} 0.9465  & \cellcolor{tabgray} 0.8373  & \cellcolor{tabgray} 0.6937  & \cellcolor{tabgray} 0.9740   & \cellcolor{tabgray} 0.9073  & \cellcolor{tabgray} 0.7919  & \cellcolor{tabgray} 0.9686  \\
    
    \bottomrule
		
	\end{tabular}
	}
}
\label{tab:comparison}
\end{SCtable*}

\begin{SCtable*}
\centering
\caption{Ablation study results on Places2, CelebA, and Dunhuang datasets. We consider three variants, \ie, predictive filtering, JPGNet with naive fusion (N.F.) defined in \reqref{eq:naivefusion}, and JPGNet with smart fusion (S.F.) defined in \reqref{eq:fusionnetwork}.}
\small
{
    {
	\begin{tabular}{l|l|ccc|ccc|c}
		
    \toprule
     & \multicolumn{1}{c|}{Datasets} & \multicolumn{3}{c|}{Places2} & \multicolumn{3}{c|}{CelebA}  & \multicolumn{1}{c}{Dunhuang} \\
    & \multicolumn{1}{c|}{Mask Ratio} &  0\%-20\% & 20\%-40\% & 40\%-60\% & 0\%-20\% & 20\%-40\% & 40\%-60\% & Default\\
    \midrule
    \multirow{6}{*}{PSNR}  
    & \cellcolor{white} Input & 16.376  & 11.271  & 9.127 & 16.259  & 11.051 & 8.887  & 15.897   \\
      & \cellcolor{white} Predictive~Filtering & 25.500  & 17.652  & 13.764 & 26.986   & 18.837  & 14.777  & 37.021  \\
    & \cellcolor{white}JPGNet-StructFlow-N.F. & 25.580  & 17.665  & 13.795 & 29.189   & 21.551  & 17.603  & 36.703  \\
    & \cellcolor{tabgray} JGPNet-StructFlow-S.F.  & \cellcolor{tabgray} 31.381    & \cellcolor{tabgray}  24.889  & \cellcolor{tabgray}  20.783  & \cellcolor{tabgray}  33.781  &  \cellcolor{tabgray}  26.501  & \cellcolor{tabgray}  21.477 &  \cellcolor{tabgray} 37.268  \\
    \cline{2-9}
    &  \cellcolor{white} JPGNet-EdgeConnect-N.F. & 25.513  & 17.663  & 13.793 & 29.315 & 21.537  & 17.489  & 37.265 \\
     &  \cellcolor{tabgray} JPGNet-EdgeConnect-S.F.  & \cellcolor{tabgray} 31.337    & \cellcolor{tabgray}  24.388  & \cellcolor{tabgray}  20.283  & \cellcolor{tabgray}  34.110  &  \cellcolor{tabgray}  26.351  & \cellcolor{tabgray}  21.195 &  \cellcolor{tabgray} 37.369 \\
    \cline{2-9}
    & \cellcolor{white} JPGNet-RFRNet-N.F. & 25.510  & 17.659  & 13.764 & 29.375   & 21.537  & 17.489   & 37.217 \\
     &  \cellcolor{tabgray} JPGNet-RFRNet-S.F.  & \cellcolor{tabgray} 30.834    & \cellcolor{tabgray}  24.056  & \cellcolor{tabgray}  19.954  & \cellcolor{tabgray}  35.898  &  \cellcolor{tabgray}  27.639  & \cellcolor{tabgray}  22.089 &  \cellcolor{tabgray} 37.646 \\
    \midrule
    \midrule
    \multirow{6}{*}{SSIM}  
    & \cellcolor{white} Input & 0.8366  & 0.6069  & 0.4165 & 0.8365  & 0.6094 & 0.4226  & 0.7661  \\
    & \cellcolor{white} Predictive~Filtering & 0.9180 & 0.7572  & 0.5913   & 0.9281  & 0.7804  & 0.6290  & 0.9692 \\
  
    & \cellcolor{white} JPGNet-StructFlow-N.F. & 0.9187 & 0.7586  & 0.5930   & 0.9338  & 0.7978  & 0.6468  & 0.9674 \\
    &  \cellcolor{tabgray} JPGNet-StructFlow-S.F.  & \cellcolor{tabgray} 0.9493    & \cellcolor{tabgray}  0.8489  & \cellcolor{tabgray}  0.7141  & \cellcolor{tabgray}  0.9640  &  \cellcolor{tabgray}  0.8875  & \cellcolor{tabgray}  0.8025 &  \cellcolor{tabgray} 0.9701  \\
    \cline{2-9}
    &  \cellcolor{white} JPGNet-EdgeConnect-N.F.  & 0.9181  & 0.7574  & 0.5902 & 0.9352   & 0.7997  & 0.6447 & 0.9701 \\
    &  \cellcolor{tabgray} JPGNet-EdgeConnect-S.F.  & \cellcolor{tabgray} 0.9516    & \cellcolor{tabgray}  0.8493  & \cellcolor{tabgray}  0.7114  & \cellcolor{tabgray}  0.9670  &  \cellcolor{tabgray}  0.8909  & \cellcolor{tabgray} 0.7725 &  \cellcolor{tabgray} 0.9718  \\
    \cline{2-9}
    & \cellcolor{white} JPGNet-RFRNet-N.F. & 0.9180  & 0.7572  & 0.5894  & 0.9358   & 0.8007  & 0.6464  & 0.9697 \\
    &  \cellcolor{tabgray} JPGNet-RFRNet-S.F.  & \cellcolor{tabgray} 0.9465    & \cellcolor{tabgray}  0.8373  & \cellcolor{tabgray}  0.6937  & \cellcolor{tabgray}  0.9740 &  \cellcolor{tabgray}  0.9073  & \cellcolor{tabgray}  0.7919 &  \cellcolor{tabgray} 0.9723 \\
    
    \bottomrule
		
	\end{tabular}
	}
}
\label{tab:ablation}
\end{SCtable*}

\begin{SCtable*}
\centering
\caption{Average Time cost (ms) per image of JPGNet and its three modules, \ie, predictive filtering~\&~uncertainty network (PFUNet), uncertainty-aware fusion network (UAFNet), and deep generative network (DGNet), on the three datasets. We also report the rate of each module w.r.t. the total cost.}
\footnotesize
{
    {
	\begin{tabular}{l|cccc|cccc|cccc}
		
    \toprule
     \multicolumn{1}{c|}{} & \multicolumn{4}{c|}{JPGNet-StructFlow} & \multicolumn{4}{c|}{JPGNet-EdgeConnect}  & \multicolumn{4}{c}{JPGNet-RFRNet} \\
    Time (ms) &  PFUNet & UAFNet & DGNet & Total & PFUNet & UAFNet & DGNet & Total & PFUNet & UAFNet & DGNet & Total\\
    \midrule
    Places2 & 4.42  & 3.33  & 30.55 & 38.30 & 4.69  & 3.40 & 15.05  & 23.14  & 3.20  & 3.10 & 15.95 & 22.25  \\
     Rate (\%) & 11.54  & 8.69  & 79.77 & 100.0   & 20.27 & 14.69 & 65.03 & 100.0  & 14.38 & 13.93 & 71.69 & 100.0  \\
    \midrule
     CelebA & 4.48   & 3.39  & 30.66 & 38.53 & 4.80  & 3.42  & 15.84  & 24.06  & 3.55  & 3.42 & 17.00 & 23.97  \\
     Rate (\%) & 11.63  & 8.80  & 79.57 & 100.0 & 19.95 & 14.21  & 65.84 &  100.0    & 14.81 & 14.27 & 70.92 & 100.0 \\
    \midrule
    Dunhuang & 4.88   & 3.51  & 31.26 & 39.65 & 4.62    & 3.37  & 14.97  & 22.96  & 3.32  & 3.33 & 16.78 & 23.43  \\
    Rate (\%) & 12.31  & 8.85  & 78.84 & 100.0 & 20.12   & 14.68  & 65.20 & 100.0 & 14.17 & 14.21 & 71.62 & 100.0 \\
    \bottomrule
	\end{tabular}
	}
}
\label{tab:time}
\end{SCtable*}

{\bf Quantitative comparison.} We first compare the three deep generative networks, \ie, StructFlow \cite{ren2019structureflow}, EdgeConnect \cite{nazeri2019edgeconnect}, and RFRNetNet \cite{li2020recurrenta}, with the respective enhanced counterparts through the proposed JPGNet in \tableref{tab:comparison}. We observe that: \ding{182} With our JPGNet, all deep generative networks are significantly improved on all three different mask ratios and three datasets, referring that our method indeed makes up drawbacks of the generative network to let the output be more identical to the ground truth. \ding{183} In terms of the time cost, as shown in \tableref{tab:time}, our method only takes slightly more time (around 8ms) than the original deep generative networks (DGNets) (See `Total' column in \tableref{tab:time}) due to the high-efficiency of predictive filtering \cite{guo2020efficientderain} and smart fusion. Overall, our method is able to enhance the generative network-based inpainting methods significantly with slight extra time cost.

{\bf Qualitative comparison.}
We further present the visualization results for the qualitative comparison in \figref{fig:viscmp}. We have the following observations: \ding{182} JPGNet is able to remove the artifacts of deep generative networks, leading to more natural inpainting results that is more identical to the ground truth. For example,  in Case2 and Case3, all of the three baseline methods show obvious artifacts, which are limited by using JPGNet. \ding{183}  JPGNet can help the deep generative networks avoid the structure distortions. For example, in the Case1, the EdgeConnect method makes the edge of steps distorted while the JPGNet-EdgeConnect recovers the arc edge of the steps. In Case4, the EdgeConnect can fill the missing pixels while generating a natural face. Nevertheless, the generated face looks much older than the real one (\ie, ground truth). In contrast, our method can produce a more identical face to the ground truth. To further validate this, we conduct a visualization comparison on Dunhuang dataset. As shown in \figref{fig:dunhuang}, all deep generative networks and their JPGNet counterparts can generate visually natural inpainting results that looks identical to the ground truth. Nevertheless, when we compare the difference maps between the completed images and the ground truth, we see that both EdgeConnect and StructFlow lead to obvious intensity variations across all pixels even most of them are not lost. In contrast, JPGNet-enhanced versions only have clear differences on missing regions, which further demonstrates that our method makes the results of deep generative models more identical to the ground truth.

\subsection{Ablation Study}

{\bf Quantitative results.} We consider three variants of our method, \ie, predictive filtering in \secref{subsec:predfilter}, naive fusion for joint predictive filtering and generative network in \secref{subsec:predict_uncertainty}, and the proposed smart fusion based on uncertainty-aware fusion network in \secref{subsec:uncerntainty}. We report the results in \tableref{tab:ablation} and have the following observations: \ding{182} Comparing with input images' PSNR and SSIM, predictive filtering presents obvious quality improvement, demonstrating the effectiveness of filtering for the image inpainting task. \ding{183} With the naive fusion strategy, the PSNR and SSIM of predictive filtering is slightly enhanced since the deep generative networks introduce more information according the whole understanding of the whole scene. \ding{184} The proposed smart fusion, \ie, UAFNet, presents significant improvements over the naive fusion on the combination of predictive filtering and deep generative networks. \ding{185} In terms of the time cost reported in \tableref{tab:time}, the PFUNet and UAFNet only takes around 4.5~ms and 3.4~ms per image on the three datasets, which is significantly less than the deep generative network (DGNet). As a result, JPGNet only leads to slight extra time costs. 

{\bf Qualitative results.} As shown in \figref{fig:fusioncmp}, the naive fusion based on the uncertainty map fills most of missing pixels but still leave some small missing regions. The UAFNet-based smart fusion can fill all missing pixels with natural appearance while recovering local structures and removing some artifacts introduced by the RFRNet. 

\section{Conclusion}\label{sec:concl}

In this paper, we have formulated the image inpainting as a mix of two problems, \ie, deep predictive image filtering and deep image generation. Predictive filtering excels at preserving local structures and removing artifacts but falls short to complete the large missing regions. The deep generative network can fill the numerous missing pixels based on the understanding of the whole scene but hardly restores the details identical to the original ones.
To make use of the best of both worlds, we propose the \textit{joint predictive filtering and generative network (JPGNet)} that contains three branches: \textit{predictive filtering~\&~uncertainty network (PFUNet)}, \textit{deep generative network}, and \textit{uncertainty-aware fusion network (UAFNet)}.
The PFUNet can adaptively predict pixel-wise kernels for filtering-based inpainting according to the input image and output an uncertainty map. This map indicates the pixels should be processed by filtering or generative networks, which is further fed to the UAFNet for a smart combination between filtering and generative results.
%
%
We have validated our method on three public datasets, \ie, Dunhuang, Places2, and CelebA, and demonstrated that our method can enhance three state-of-the-art generative methods (\ie, StructFlow, EdgeConnect, and RFRNet) significantly with slightly extra time cost.

\begin{acks}
This work has partially been sponsored by the National Science Foundation of China (No. U1803264 and No. 61672376). 
It was also supported by the Natural Science Foundation of Tianjin under Grant No. 20JCQNJC00720, the National Research Foundation, Singapore under its the AI Singapore Programme (AISG2-RP-2020-019), the National Research Foundation, Prime Ministers Office, 
Singapore under its National Cybersecurity R\&D Program (No. NRF2018NCR-NCR005-0001), NRF Investigatorship NRFI06-2020-0001, 
the National Research Foundation through its National Satellite of Excellence in Trustworthy Software
Systems (NSOE-TSS) project under the National Cybersecurity R\&D (NCR)
Grant (No.~NRF2018NCR-NSOE003-0001). 
We gratefully acknowledge the support of NVIDIA AI Tech Center (NVAITC) to our research.
\end{acks}

\bibliographystyle{ACM-Reference-Format}
\balance
\bibliography{ref}

\appendix

\end{document}